\documentclass[letterpaper]{article} 
\usepackage{aaai24}  
\usepackage{times}  
\usepackage{helvet}  
\usepackage{courier}  
\usepackage[hyphens]{url}  
\usepackage{graphicx} 
\usepackage{natbib}  
\usepackage{caption} 
\usepackage{algorithm}
\usepackage{algorithmic}

\usepackage{newfloat}
\usepackage{listings}
\DeclareCaptionStyle{ruled}{labelfont=normalfont,labelsep=colon,strut=off} 
\floatstyle{ruled}
\newfloat{listing}{tb}{lst}{}
\floatname{listing}{Listing}
\usepackage{xspace}
\newcommand{\xhdr}[1]{\noindent{{\bf #1.}}} 
\usepackage{amsthm} 
\usepackage{amsmath}
\usepackage{amssymb}
\newtheorem*{problem*}{Problem} 
\usepackage{array}  
\usepackage{booktabs}
\usepackage{longtable}
\newcommand{\remove}[1]{} 
\usepackage{multirow}
\usepackage{tabularx}
\usepackage{booktabs}
\usepackage{etoolbox}
\usepackage{paralist}
\usepackage{subcaption} 
\usepackage[table]{xcolor}
\usepackage{enumerate}
\usepackage[shortlabels]{enumitem}
\usepackage{soul} 
\newcommand{\docs}{\ensuremath{\mathcal{D}}}

\usepackage{lipsum}

\newcommand{\q}{$q$\xspace}

\newcommand{\filteringscore}{bio-inspiration score\xspace} 

\newcommand{\BARcode}{\textsc{BARcode}\xspace}

\title{Imitation of Life: A Search Engine for Biologically Inspired Design}
\author {
    Hen Emuna\textsuperscript{\rm 1},
    Nadav Borenstein\textsuperscript{\rm 2},
    Xin Qian\textsuperscript{\rm 3},
    Hyeonsu Kang\textsuperscript{\rm 4},\\
    Joel Chan\textsuperscript{\rm 3},
    Aniket Kittur\textsuperscript{\rm 4},
    Dafna Shahaf\textsuperscript{\rm 1}
}
\affiliations {
    \textsuperscript{\rm 1}The Hebrew University of Jerusalem\\
    \textsuperscript{\rm 2}University of Copenhagen\\
    \textsuperscript{\rm 3}University of Maryland\\
    \textsuperscript{\rm 4}Carnegie Mellon University\\

    hen.emuna@mail.huji.ac.il, nadav.borenstein@di.ku.dk, xinq@umd.edu, hyeonsuk@cs.cmu.edu, joelchan@umd.edu, nkittur@cs.cmu.edu, dshahaf@cs.huji.ac.il  
}

\begin{document}

\maketitle

\begin{abstract}
Biologically Inspired Design (BID), or Biomimicry, is a problem-solving methodology that applies analogies from nature to solve engineering challenges. For example, Speedo engineers designed swimsuits based on shark skin. Finding relevant biological solutions for real-world problems poses significant challenges, both due to the limited biological knowledge engineers and designers typically possess and to the limited BID resources. Existing BID datasets are hand-curated and small, and scaling them up requires costly human annotations.

In this paper, we introduce \BARcode (Biological Analogy Retriever), a search engine for automatically mining bio-inspirations from the web at scale. Using advances in natural language understanding and data programming, \BARcode identifies potential inspirations for engineering challenges. Our experiments demonstrate that \BARcode can retrieve inspirations that are valuable to engineers and designers tackling real-world problems, as well as recover famous historical BID examples. We release data and code; we view \BARcode as a step towards addressing the challenges that have historically hindered the practical application of BID to engineering innovation.
\end{abstract}

\section{Introduction}
Nature is a rich source of inspiration for creative problem-solving. For example, engineers addressing the bullet train's sonic boom problem drew inspiration from kingfishers, who dive without splashing water~\cite{bullettrain1}. By redesigning the train's front after a kingfisher's beak, they reduced booms and improved speed while conserving energy. 

Biologically Inspired Design (BID), or biomimicry, is a design approach that derives solutions by drawing analogies to nature's strategies and systems~\cite{fu_bio-inspired_2014,lurie2014product, dumanli2016recent, pawlyn2019biomimicry, san2020review}. 

In the past two decades, there has been a growing interest in BID~\cite{biomimetics8010107}. However, despite its potential, it is still not widely applied. One reason might be that finding relevant biological solutions is time-consuming due to limited biological knowledge among designers and engineers~\cite{vattam2011foraging}. Analogies from biology often rely on serendipity~\cite{Vattam2013SeekingBO} -- in the bullet train case, one of the engineers was an amateur birdwatcher~\cite{bulletTrain}. Moreover, while biological knowledge is abundant, bio-inspirational information is scarce. An audit of potential inspirations, involving heuristically filtered scientific articles from PubMed, revealed that only 3.36\% (from a sample of 3416) provided insights for design problems~\cite{bioscrabble}.

Several hand-curated databases of BID exist. AskNature~\cite{AskNature} is an online dataset of biological phenomena, categorized by functional keywords. Design by Analogy to Nature Engine (DANE) provides a library of biological descriptions represented in the Structure-Behavior-Function (SBF) framework~\cite{DANE, goel2009structure}. However, the hand-curation of these databases is an extremely time-consuming and costly process; it takes 40–100 hours to describe a single biological organism in their framework. Thus, there is a need for automatic, \emph{scalable} methods to expand the coverage of these datasets~\cite{vandevenne_seabird:_2016}.

\begin{figure*}[ht]
\centering
\includegraphics[width=0.95\textwidth]{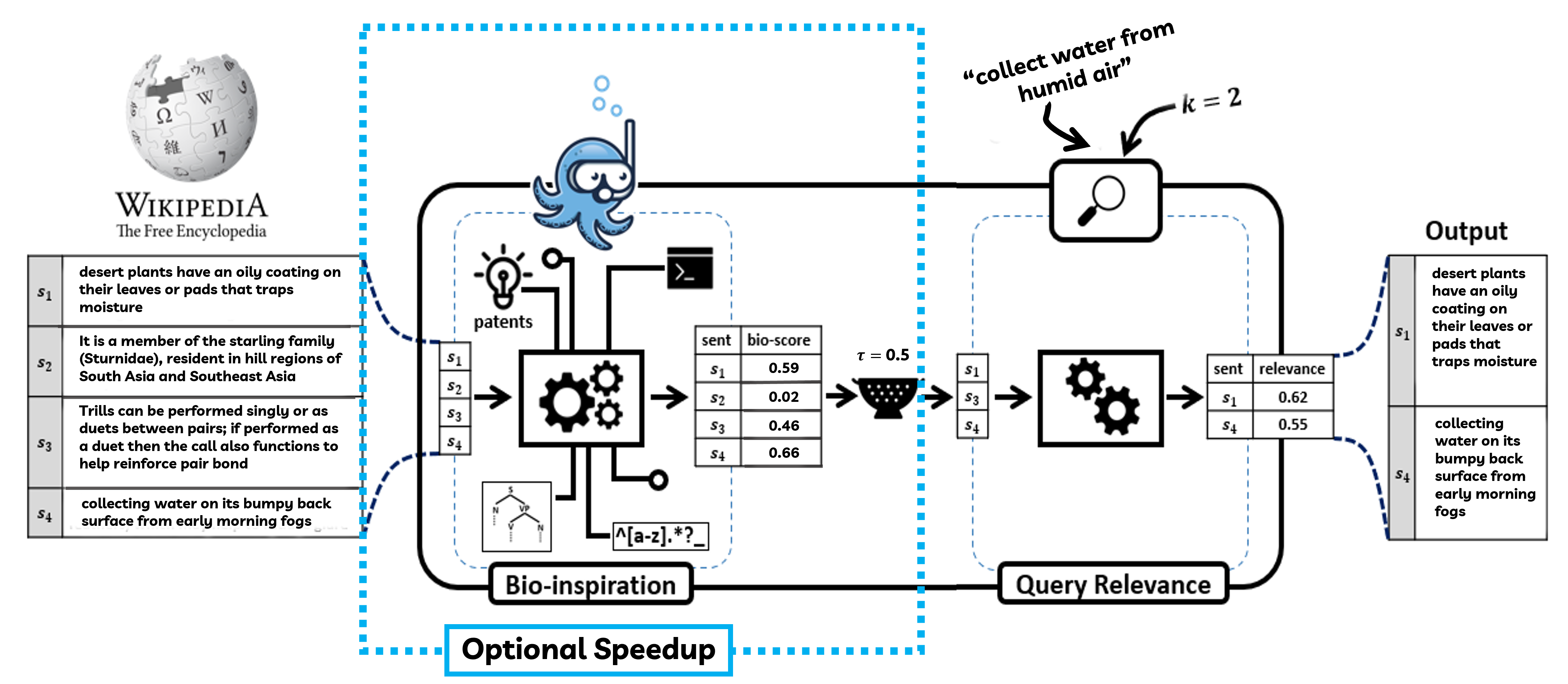}
\caption{An overview of \BARcode. The bio-inspiration module, powered by Snorkel, processes a large corpus of biological documents, like Wikipedia, and generates a bio-inspiration score for each sentence in the corpus. The score measures the likelihood that the sentence describes a biological solution to a problem and is generated using linguistic labeling functions and external data sources like patent database. The query-relevance module then filters low-scoring sentences and, given a query, returns a ranked list of length k of relevant candidate sentences sorted by their relevance to the query.}
\label{fig:system_overview}
\end{figure*}

More recently, Zhao et al. demonstrated how crowdsourcing annotations of scientific articles containing biomimicry-related content can facilitate a supervised learning approach to classifying scientific articles~\cite{IBM, biologue}. However, generating labels for a large dataset can again be expensive and difficult.

Another work relies on hard-coded rules to seek bio-inspiration~\cite{Cheong2014}. However, the set of rules is small and limited.
Furthermore, they match only the identified causal-function verbs, which are often too broad. For instance, a query like ``reduce glare'' only matches on ``reduce'', possibly leading to irrelevant matches that do not refer to ``glare''.

In this work, we introduce \BARcode (Biological Analogy Retriever), a \textit{scalable} and \textit{automatic} end-to-end system for mining bio-inspiration from unstructured data written in natural language. A recent work in analogy mining has shown how mining functional aspects such as the purposes and mechanisms of products can support the search and discovery of products that match on an abstract level \cite{hope2022scaling}.
We incorporate recent advances in natural language understanding and data programming to extract functional aspects relevant to our settings. We show our system {\bf retrieves valuable bio-inspirations} for human design problems, both in a {\bf retrieval experiment} and with {\bf real experts} looking for inspiration to solve their own research problems. We also demonstrate how \BARcode can {\bf recover famous historical biomimicry examples}.
We release data and code at \url{https://github.com/emunatool/BARcode-BioInspired-Search}.

We hope \BARcode will help to scale existing bio-inspiration datasets (e.g., AskNature) and aid designers and engineers in addressing real-world problems.

\section{Problem Formulation}
\label{sec:Problem Formulation}
Given a corpus of documents $\docs$ written in a natural language and a query \textit{q} describing a challenge (e.g., ``collect water from humid air''), our goal is to return a ranked list of sentences from the corpus containing possible sources of inspiration. Specifically, we are looking for sentences describing an organism facing a similar \textit{{challenge}}, and potentially the \textit{{strategy}} it uses to overcome it. For example, the sentence  ``Desert plants have an oily coating on their leaves or pads that traps moisture, thereby reducing water loss'' includes a {challenge} (trapping moisture/reducing water loss) and hints at a {strategy} (oily coating). 

In our formulation, we are inspired by a recent work in analogy mining that demonstrated how mining functional aspects (such as the purposes and mechanisms of products) can support the search for analogous products that match on an abstract level \cite{hope2022scaling}.
We adapt this idea into our setting, focusing on the aspects of challenge and strategy. 

\BARcode can work with any corpus containing biological knowledge. For this paper, we focus on \textit{Wikipedia} because of its accessibility and broad coverage of diverse biological topics~\cite{WikiStat}. 
The corpus contains 780,949 sentences, collected from 27,640 articles listed under the category ``Articles with 'species' microformats''\footnote{We used spaCy's NLP engine \cite{spacy2} to process the articles.}.

\section{Mining Biological Analogies}
We start by describing our algorithm (Section \ref{sec:match}), and then discuss speedups in Section \ref{sec:filter}. An overview of our system can be seen in Figure \ref{fig:system_overview}. Our algorithm consists of two phases: First, an optional speedup phase to {identify sentences that are likely to contain a challenge an organism faces (and potentially a coping strategy)}. Next, we {\bf match and rank} the most relevant sentences to \q.

\subsection{Calculating Relevance Score}
\label{sec:match}
Here, our goal is to identify the most relevant sentences that would provide useful bio-inspiration for addressing the user's challenge. 
Our system is designed to handle short queries describing a specific challenge, e.g.,  ``collect water from humid air''. As sentences from the corpus can be long and convoluted, we start by extracting candidate phrases from the sentences in the form of verb-object pairs (e.g., ``trap moisture''). 

\begin{table*}[t!]
\centering
  \begin{tabular}{ p{2cm} p{11cm} p{1.5cm}}
    \toprule
    Organism   &  Sentence    &  Extraction Method\\ \hline
    \midrule
    \multicolumn{3}{c}{\textbf{prevent sinking}} \\ \hline
    Ctenophora & If they enter less dense brackish water, the ciliary rosettes in the body cavity may pump this into the mesoglea to increase its bulk and decrease its density, to \textbf{avoid sinking} & QA-SRL\\
     Barychelidae & Others can \textbf{avoid drowning} by trapping air bubbles within the hairs covering their bodies  & QA-SRL\\
     Pelican  & The air sacs serve to \textbf{keep} the pelican remarkably \textbf{buoyant} in the water and may also cushion the impact of the pelican's body on the water surface when they dive from flight into water to catch fish    & DEP\\
     Cephalopod  & Other cephalopods use ammonium in a similar way, storing the ions as ammonium chloride to reduce their overall density and \textbf{increase buoyancy}    & DEP\\
      \hline
   
    \multicolumn{3}{c}{\textbf{collect water from humid air}} \\ \hline
     Stenocara Gracilipes & Facing into the breeze, with its body angled at 45 degrees, the beetle \textbf{catches fog droplets} on its hardened wings & Both\\
     Yucca & Some desert plants have an oily coating on their leaves or pads that \textbf{traps moisture}, thereby reducing water loss  & QA-SRL\\
     Kangaroo rat  & To reduce loss of moisture through respiration when sleeping, a kangaroo rat buries its nose in its fur to \textbf{accumulate a small pocket of moist air}   & QA-SRL\\
  \bottomrule
\end{tabular}
\caption{Selected sentences from the Top 15 retrieved by \BARcode for two queries. The phrases that are used to match to the query are highlighted in bold.}
\label{tab:runing_example}
\end{table*}

\subsubsection{Extracting Candidate Phrases}
To extract verb-object pairs, we use two complementary techniques. The first technique employs Semantic Role Labeling (SRL) \cite{SRL}, specifically we use QA-SRL \cite{DBLP:journals/corr/abs-1805-05377}. QA-SRL is a model that takes a sentence as input and generates question-answer pairs related to the verbs\footnote{We omit 'When' and 'Who' questions and those not ending with a verb. Based on experimentation with the QA-SRL model, these question types usually don't align with challenge-related verb-object pairs.}. For example, applying QA-SRL on the sentence from Section \ref{sec:Problem Formulation}: Q: `What does something trap?', A:`moisture'. We then convert each question-answer pair into the format '[verb] [answer]'\footnote{[verb] represents the lemma form of the main verb in the question, and [answer] represents the span of the answer} (e.g., `trap moisture').

However, QA-SRL has limitations. It fails to extract certain challenge-related phrases, especially those with auxiliary verbs (e.g., ``keep buoyant''); it tends to produce verbose phrases (``store ions to reduce overall density and increase buoyancy''). To address these limitations, we employ a second technique, pattern-matching on dependency trees with spaCy~\cite{spacy2}. Dependency parsing offers greater flexibility in extracting verb-object pairs, allowing connections between non-adjacent verbs and objects. Through experimentation, we have found 10 patterns (see Appendix A.1).
By employing these two techniques, we extracted 2,829,204 candidate phrases from sentences.

\subsubsection{Matching and Ranking}
\label{matching_ranking}
Next, we employ a two-step approach to assess the similarity between these candidate phrases and the query. We employed SBERT, designed for semantic search~\cite{Sentence_transformer}\footnote{https://huggingface.co/sentence-transformers/multi-qa-mpnet-base-dot-v1}, and DeBERTa~\cite{he2020deberta}\footnote{https://huggingface.co/cross-encoder/nli-deberta-v3-base} as the inference model.

Sentence embedding models, like SBERT, effectively capture semantic meaning, but they have trouble handling antonyms (e.g., ``increase'' and ``decrease'' might be considered similar)~\cite{Polarity_antonyms, antonyms_space_model}, which is problematic for our use case. Additionally, these models struggle with causality and common sense, causing them to overlook relevant matches (e.g., make the connection between ``stay moist'' and ``reduce water loss'').

To address these limitations, we add to our pipeline a Natural language inference (NLI) model~\cite{NLI}. NLI is the task of determining whether a hypothesis is true (entailment), false (contradiction), or undetermined (neutral) given a premise, and can address the shortcoming of the embedding similarity techniques~\cite{vahtola-etal-2022-easy, lobue-yates-2011-types}. 

We chose DeBERTa, which has shown good performance on prominent NLI benchmarks. Candidate phrases form the premise and the query forms the hypothesis, resulting in three scores for every pair (entailment, neutral, and contradiction).
We note that DeBERTa indeed manages to identify that ``stay moist'' implies ``reduce water loss'' and that ``increase water loss'' contradicts ``reduce water loss''. 

Using the inference model is computationally costly. Thus, we restrict ourselves to the top 4,000 phrases according to semantic similarity (SBERT). Looking at the data, 4000 seems to provide a large safety margin, so we are not likely to filter out any promising candidate.

To create a single weighted score, we train a simple classifier to classify pairs as \textit{relevant} or \textit{irrelevant} based on the NLI scores and SBERT score (See Appendix A.1 for details).

We retrieve the top sentences according to our classifier. 
In Table \ref{tab:runing_example}, we provide examples of retrieved sentences for two queries: ``prevent sinking'', and ``collect water from humid air''.

\subsection{Speeding up the algorithm}
\label{sec:filter}

We built \BARcode to work with any corpus
containing biological knowledge. However, sentences that could serve as biomimetic inspiration are typically very rare in such corpora. When the corpus is very large, running the algorithm could become computationally expensive. Thus, in this section, we suggest filtering heuristics to narrow down the sentences we consider as candidates to match with a query.

To implement the filtering algorithm, we leverage recent advancements in \textbf{data-programming}. Data-programming relies on three key elements: 1) extracting \textit{candidates} for labeling, 2) generating \textit{labeling functions} that express domain heuristics (distant supervision), and 3) using a \textit{generative model} to handle noise and correlations among labeling functions, resulting in probabilistic labels for all candidates. 

We use {Snorkel} \cite{Snorkel}, defining a candidate as three text spans (\textit{Strategy}, \textit{Solver}, and \textit{Problem}). Note that sentences might include several candidates.

We employed ten clausal patterns to detect candidates. Seven patterns were adapted from \citet{Cheong2014} (e.g., ``the \textit{Strategy} serves as a clausal subject for the \textit{Problem} and is in gerund form''). After experimenting with the data, we added three more patterns: (1) \textit{Problem} functions as an adverbial clause modifier, modifying the \textit{Strategy}, (2) \textit{Problem} is the main clause head of a relative clause, in which the \textit{Strategy} appears in the main clause, and (3) \textit{Problem} is the main clause head of an adjectival clause, where the \textit{Strategy} serves as the sentence's root. Out of 494,640 dataset entries, approximately 234,000 contained one or more clausal patterns and were used as candidates.

\xhdr{Defining Labeling Functions (LF)}
We defined labeling functions (LFs) to assign each candidate to one of three labels: contains bio-inspiration, does not contain bio-inspiration, or unknown.

Some LFs look for cues that a sentence contains bio-inspiration:
\begin{itemize}
    \setlength\itemsep{0em} 
    \item \textbf{Keywords:} The sentence contains the word ``adaptation'' or one of its conjugations. We notice that almost all the sentences that contain this word indeed describe an interesting analogy. For example,  \textit{``The eyes appear to be narrowly open due to the lowered upper eyelid, probably an \textbf{adaptation} to shield the eyes from the sun's glare''}.

    \item \textbf{Distant Supervision - Patents:} The sentence contains a description of a known problem. Candidates that contain a known problem are also likely to contain a solution to this problem. We used a dataset of {\bf patents} to construct a list of known problems (using what one might call ``cross-corpus distant supervision''): First, we downloaded 20 million sentences from the ``Claims'' section of a random set of patents. Claims describe what the patent is supposed to do. Then, we extracted from this section strings in the form of ``for [verb]-ing [noun]'' using regular expressions and POS (Part Of Speech) tags. Coming from the claims section, those strings are likely to address the problem that the patent was designed to solve. For example, \textit{``A surface cleaning apparatus ... comprising a second collecting apparatus ... \textbf{for collecting liquid} from the surface''}.
    
    We listed the $2,000$ most common verb-noun pairs and used them as our known problems list. This list contains pairs like ``tilt movement'',  ``extract signal'' and ``sense light''. We say that a candidate describes a known problem if it contains such a noun-verb pair, which are close to one another in the sentence. Note that many of the problems coming from patents are unlikely to be solved by animals (e.g., ``encode information'').

    \item \textbf{Auxiliary Verbs:} The sentence contains an auxiliary verb like ``allow'', ``help'' or ``enable''. Those verbs exist in many relevant candidates that describe a biological trait, that is used by an organism to solve a problem. The problem, in this case, will appear right after the auxiliary verb, as seen in the following sentence: \textit{The webbing between the toes increases the area of the foot and \textbf{helps} propel the frog powerfully through the water''}.
\end{itemize}

{Other LFs look for cues that a sentence is \emph{unrelated} to biological strategies:}
\begin{itemize}
    \setlength\itemsep{0em} 
    \item \textbf{Verbs with No Biology Relation:} The main verb of the sentence is not biology-related. Among those, we can find verbs such as ``dance'', ``read'', ``explain'', and ``discover''. This list of verbs was hand-crafted from the 200 most common verbs in the English language and contains 36 entries.
    \item \textbf{Unlikely Entities:} The sentence contains a component that is not expected to be found in functional sentences. For example, if the sentence contains the name of a person or organization, a date, a pronoun, or a symbol (such as \&, @, or \$), it is unlikely that it describes a biological solution to a problem.
\end{itemize}

Next, Snorkel predicts the likelihood that a candidate is bio-inspirational. Note that this process is performed as a pre-processing step and does not depend on the query.

\begin{table}[htb]
  \centering
  \begin{tabular}{p{6cm}p{1.4cm}}
    \toprule
    \textbf{Candidate sentence} & \textbf{Snorkel's Score} \\
    \midrule
    \rowcolor{green!50!white}
    \textit{Peregrine falcon} --- The air pressure from such a dive could possibly damage a bird's lungs, but \textbf{small bony tubercles} on a falcon's nostrils guide the powerful airflow away from the nostrils, enabling the bird to breathe more easily while diving by \textbf{reducing the change in air pressure} & \textbf{0.935} \\ \hline
    \rowcolor{green!35!white}
    \textit{Isopoda} --- The dorsal (upper) \textbf{surface} of the animal is \textbf{covered by a series of overlapping, articulated plates} which \textbf{give protection while also providing flexibility} & \textbf{0.802} \\ \hline
    \rowcolor{green!20!white}
    \textit{Yucca} --- Some desert plants have an \textbf{oily coating} on their leaves or pads that \textbf{traps moisture}, thereby reducing water loss & \textbf{0.598} \\ \hline
    \rowcolor{red!10!white}
    \textit{Pigeon guillemot} --- Trills can be performed singly or as duets between pairs; if performed as a duet then the call also functions to help reinforce pair bond & 0.469 \\ \hline
    \rowcolor{red!35!white}
    \textit{Morgan horse} --- By the 1870s, however, longer-legged horses came into fashion, and Morgan horses were crossed with those of other breeds & 0.213 \\ \hline
    \rowcolor{red!50!white}
    \textit{Common hill myna} --- It is a member of the starling family (Sturnidae), resident in hill regions of South Asia and Southeast Asia & 0.0002 \\
    \bottomrule
  \end{tabular}
\caption{Examples of Snorkel's score for the filtering of the dataset.  \textbf{Challenge} and \textbf{strategy} are highlighted in bold. }
\label{tab:filtering_score}
\end{table}

See Table \ref{tab:filtering_score} for some of Snorkel's scores.
We set the filtering threshold to $\tau=0.5$ (after an examination of the data, this seems to nicely balance precision and recall), leaving us with 23,553 sentences (3\% of the total data). In the evaluation section we show that running on this much-smaller set of sentences does not harm performance much.

\section{Evaluation}
In this section, we report our evaluation of the performance of \BARcode's \filteringscore module, as well as an end-to-end evaluation of the system's ability to retrieve biomimetic inspirations.

Our research questions are as follows:
\begin{itemize}
    \setlength\itemsep{0em} 
    \item \textbf{RQ1:} Can our approach provide inspiration from nature to solve human design problems?   
    \item \textbf{RQ2:} Is our algorithm robust to query variations?
    \item \textbf{RQ3:} Does the bio-inspiration module predict whether a sentence contains a challenge? 
\end{itemize}

\xhdr{A Note on Large Language Models}
Large language models have recently gained immense popularity, achieving state-of-the-art results across a variety of downstream tasks. We have initially experimented with applying ChatGPT (\url{https://openai.com/chatgpt}) to our problem, but quickly found out that it 1) tends to hallucinate facts that we could not corroborate (``Termites build mounds that capture and collect moisture from the air through a process known as passive condensation''), making it hard to trust, and 2) has limited coverage (and when pressed to come up with more organisms, simply repeats the first ones). Thus, we decided that an information-retrieval setting that covers the entire Wikipedia and can point our users to the source of each claim is better suited for our use case.

\subsection{Evaluating Retrieval Quality}
 In this section, we evaluate the retrieval quality of our system to tackle (\textbf{RQ1}) –- can our algorithm provide relevant inspirations to solve human design problems?

\subsubsection{Reality check: Famous Historical Examples}
\label{subsec:famous_biomimicry_examples}
To assess our system's retrieval quality, we first try it on famous historical biomimicry examples. We scoured the 'INNOVATIONS' section of AskNature, which outlines nature-inspired breakthroughs, and sampled five examples to formulate queries and search for them. The results were:
\begin{compactitem}
    \item Capturing cooling tower plumes to reduce water usage, inspired by Namibian desert beetles. Ranked 8 for the query ``collect water from humid air''. 
    \item Residue-free adhesion technology, inspired by gecko feet. Ranked 9 for the query ``provide adhesion''.
    \item Speedo's swimsuit for faster swimming, inspired by shark skin. Ranked 10 for the query ``reduce fluid drag''.
    \item Passively cooled buildings, inspired by termite mounds. Ranked 14 for the query ``provide self-regulating ventilation system''.
    \item Train’s front inspired by a kingfisher’s beak, not found. We checked the kingfisher's Wikipedia page, and it does not mention the beak's hydrodynamic structure.
\end{compactitem} 

We note that famous historical examples might be easier to find, as Wikipedia is more likely to contain a sentence about them; thus, we consider this a reality check and move on to further evaluation of the retrieval quality.

\subsubsection{Retrieval Experiment: Collecting queries}
To collect relevant innovation challenges, we used titles from AskNature's ``Biological Strategies'' section. These titles outline how organisms address challenges (e.g., ``Adaptive camouflage helps blend into the environment''). We converted these titles into query format (verb-object pairs, e.g., ``blend into environment''). This resulted in 24 queries.

To assess the robustness of our system (\textbf{RQ2}) (Section \ref{subsec:Robustness_to_Query_Variations}), we generated more technical variations for 18 queries
(e.g., ``repel water''/``prevent water absorption'').

\xhdr{Baseline} 
As we could not access the code for the work of \citet{Cheong2014}, we compared our system with a search engine (which is the most likely tool to be used by people looking for biomimetic inspiration today). We used Elasticsearch (v7.13.2), a widely used search engine built on top of the Lucene library. We used Elasticsearch's default relevance scoring function. Both for our algorithm and for the baseline, we consider two settings:

\begin{compactitem}
    \item \textbf{Entire data:} Used all sentences in the corpus.  
    \item \textbf{Filtered data:} Only sentences passing the threshold $\tau=0.5$, corresponding to aggressive pruning of the data (keeping 3\% of the entire data). See Section \ref{sec:filter}.
\end{compactitem}

\xhdr{Experiment design}
We collected the top 15 results per query from \BARcode and the baseline for both settings, resulting in 1673 unique sentences. 
We used crowdsourcing to determine whether each sentence contains a \textit{challenge} and/or \textit{strategy}.
Annotators were presented with an innovation challenge (query) along with a sentence about an organism (see Appendix A.3). 
Annotators were asked to assess whether the sentence indicates: (a) the organism faces a similar challenge, (b) provides a strategy to tackle the challenge, (c) both, or (d) is irrelevant. 

We hired 29 Amazon Mechanical Turk (AMT) crowd workers who passed a qualification test. They received \$0.11 per task.
Each sentence was annotated by 3 workers, and the presence of a challenge and/or strategy was determined by a majority vote. 

The annotators showed moderate agreement, as indicated by Fleiss' Kappa scores of 0.42 for challenge and 0.55 for strategy evaluation. This might indicate the difficulty of the task: sentences sometimes lack context, and challenges and strategies are sometimes implicit. We note that some annotators preferred to mark only the more explicit sentences, while others did not show such preference.

\xhdr{Results}
Table \ref{tab:precision_ndcg_algorithm_baselines_comparison_all_queries} compares {\BARcode} with the baseline, both on the entire data and filtered data, using two information retrieval metrics: \textit{Precision @k} and \textit{Normalized Discounted Cumulative Gain (NDCG)@k} at two cut-offs (k=7, k=15). For NDCG, we use binary relevance.

Our system outperforms the baseline in all metrics. Most importantly, on the entire data, 72\% of the top 7 sentences and 69\% of the top 15 sentences retrieved by \BARcode contain a strategy that could be inspirational for solving the challenge (which is quite high in absolute terms). In comparison, the baseline only yields 36\% for the top 7 and 29\% for the top 15.  

The situation is similar for sentences that contain a challenge, with our system outperforming the baseline. We note that the absolute percentage of those sentences is lower (which surprised us). Looking at the data, this seems to stem from the fact that annotators tended not to mark sentences where the challenge was somewhat implicit. 

On the filtered data, our system again beats the baseline on all metrics. 
There is a trade-off between speed (filtered data is only 3\% of the entire data) and performance. For both \BARcode and the baseline, filtering does not hurt performance much for the top 7 but starts to be more pronounced at the top 15. We believe that a less aggressive threshold would have narrowed the gap; we leave the problems of finding automated ways to pick a threshold for future work.

We used the Mann-Whitney U test to assess the statistical significance of differences in precision between \BARcode and baseline on the entire data, and also between \BARcode on the entire data and the filtered data. This analysis was conducted separately for \textit{challenge} and \textit{strategy} at two cut-off points: k=7 and k=15, covering all queries (with $\alpha$ of 0.05 before Bonferroni correction). Comparing \BARcode to the baseline, all tests were statistically significant (p-values: 2.26e-7 for k=7, 2e-10 for k=15 for strategy, and 5.96e-6 for k=7, 2.39e-8 for k=15 for challenge). Differences between \BARcode on the entire data and filtered data were not significant, except for strategy at k=15 (p-value=3.88e-3). This supports our observations about the utility of filtering.

\begin{table}[t!]
\begin{tabular}{p{1.2cm}ccc}
 & {\textbf{Method}} & \textbf{P} & \textbf{NDCG}  \\ \cmidrule(l){2-4} 
\multirow{6}{*} {\textbf{Challenge}} & {Baseline Filtered @7}   & 0.272  & 0.554   \\ & Baseline Entire @7   & 0.282  & 0.536  \\ &  {\BARcode Filtered @7}   & \textbf{0.541}  & \textbf{0.736} \\ &  \BARcode Entire @7  & \textbf{0.565}     & \textbf{0.787} \\  \cmidrule(l){2-4} & Baseline Filtered @15  & 0.224  & 0.557  \\
& Baseline Entire @15  & 0.24   & 0.554  \\ & \BARcode Filtered @15& \textbf{0.435} & \textbf{0.745} 
\\ &   \BARcode Entire @15  & \textbf{0.541} & \textbf{0.798} 
\\  \midrule\midrule \multirow{6}{*}{\textbf{Strategy}} & Baseline Filtered @7  & 0.286   & 0.565  \\  & Baseline Entire @7  & 0.357 & 0.647 \\ &  \BARcode Filtered @7  & \textbf{0.568}     & \textbf{0.763} \\ &  \BARcode Entire @7  & \textbf{0.718}  & \textbf{0.884} \\ \cmidrule(l){2-4} & {Baseline Filtered @15}  & 0.252  & 0.579  \\ & Baseline Entire @15 & 0.295  & 0.658  \\ &  \BARcode Filtered @15 & \textbf{0.514} & \textbf{0.772}\\ &  \BARcode Entire @15 & \textbf{0.694} & \textbf{0.877}    
\end{tabular}
\caption{Precision (P)@k and NDCG@k for all queries.
Both \BARcode outperforms the  baseline in every metric, both on the entire data and on filtered data. Filtering does not hurt performance much for the top 7, but starts to be more pronounced at the top 15.}
\label{tab:precision_ndcg_algorithm_baselines_comparison_all_queries}
\end{table}

\begin{figure}[t!]
     \centering
     \begin{subfigure}{\columnwidth}
         \centering
         \includegraphics[width=\columnwidth]{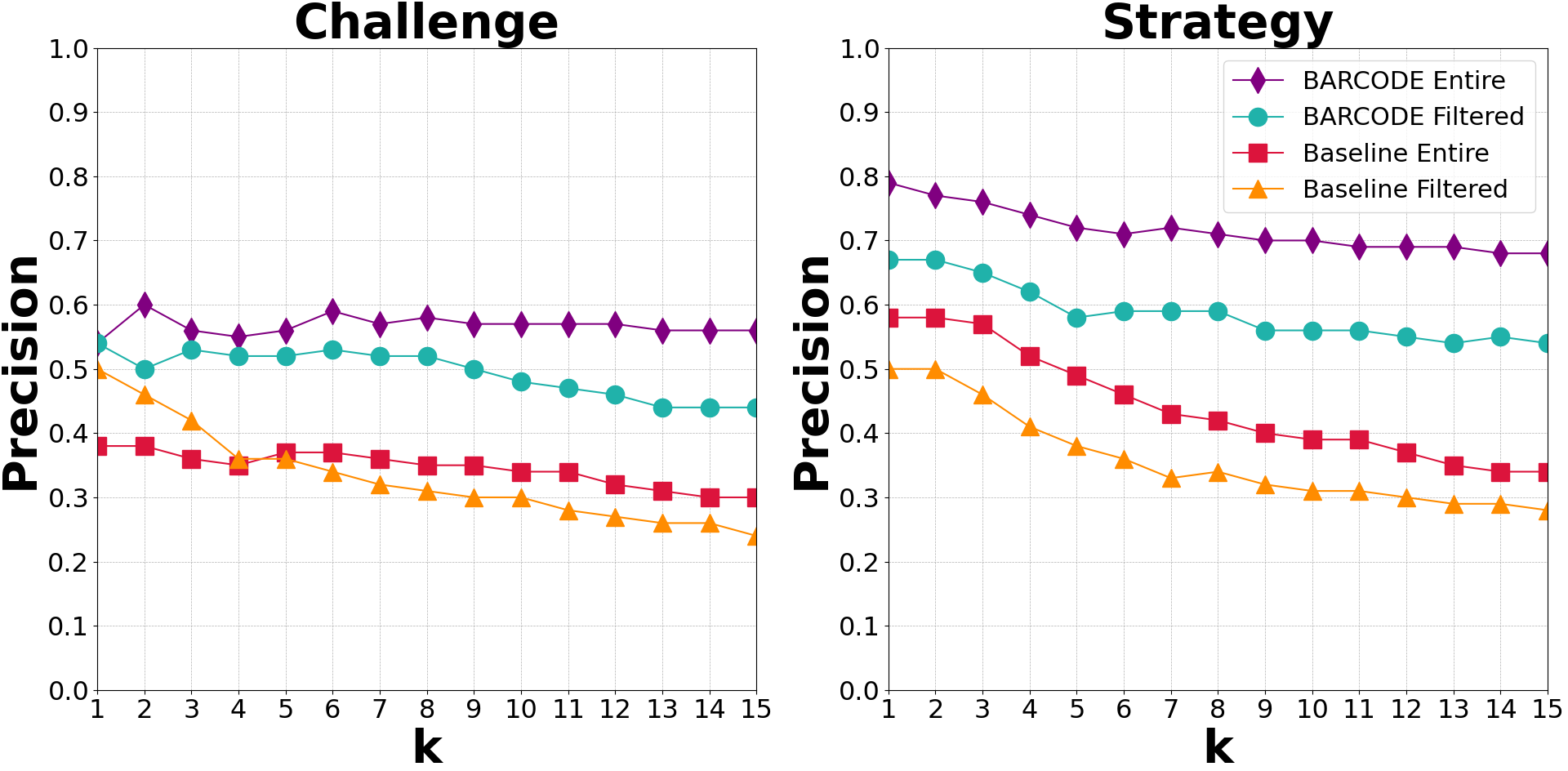}
         \caption{Ask Nature Queries}
         \label{fig:ask_nature_precision}
     \end{subfigure}
     \begin{subfigure}{\columnwidth}
         \centering
         \includegraphics[width=\columnwidth]{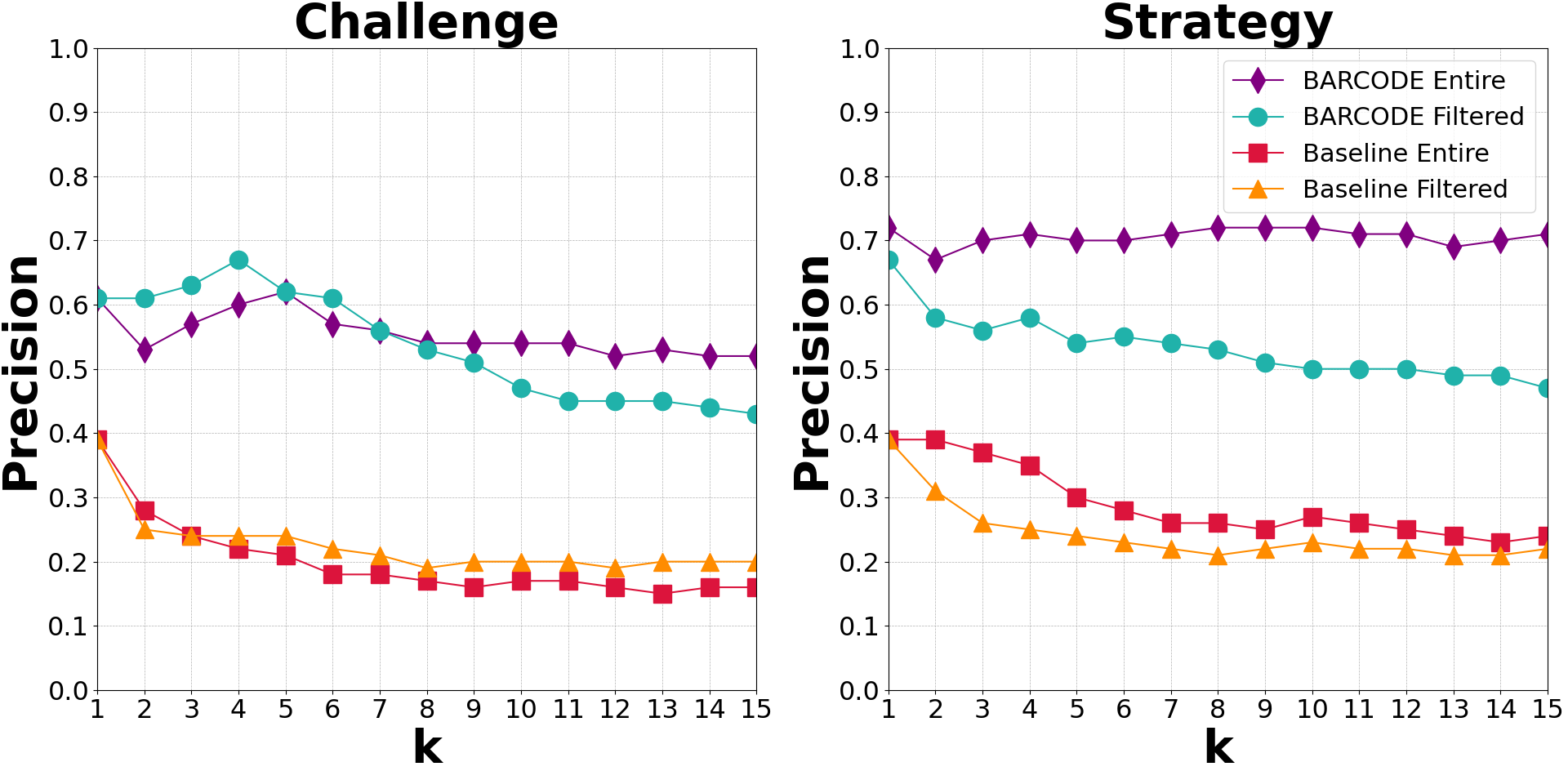}
         \caption{Paraphrased Queries}
         \label{fig:paraphrases_precision}
     \end{subfigure}
        \caption{Precision @k across all queries. \textbf{Top}: AskNature queries. \textbf{Bottom}: Paraphrased queries. \textbf{Left}: sentences containing challenges similar to the query. \textbf{Right}: sentences containing strategies to tackle the challenge in the query. \BARcode outperforms the baseline both on the entire data and filtered data. The gap in performance is wider on the paraphrased queries, perhaps due to our NLI model.}
\label{fig:precision_graph}
\end{figure}

Next, we take a look at query paraphrases.
Figure \ref{fig:precision_graph} shows Precision@k results for AskNature queries and their variations. Notably, the gap between \BARcode and the baseline is greater for the paraphrases. 
We hypothesize that this happens because the paraphrases use more technical terms than the original ones, and also sometimes require commonsense to understand the connection, which might be captured by our NLI component (but not by a regular search engine).

\xhdr{Error Analysis}
We examined the 10 queries achieving the worst performance, yielding 106 sentences annotated as irrelevant. Among these, 29\% were due to annotators overlooking challenge/strategy within sentences, often when it is implicit. 
In 39\% of cases, the matched candidate phrase did not capture the intended challenge, like ``detect carbon dioxide'' being matched to the query ``capture carbon dioxide''. In 27\% of cases, the matched candidate phrase was not actually about the organism. 
In 5\% of the cases, the sentences lack bio-inspirational content.

\smallskip

To conclude, \BARcode\ {\bf outperforms the baseline} across the board, and can {\bf provide useful inspirations} from nature to a large variety of queries (see also experiments with experts in Section \ref{sec:ideation}). Our filtering heuristics are able to drastically reduce the size of the data before performance starts to decline.

\subsection{Robustness to Query Variations}
\label{subsec:Robustness_to_Query_Variations}
We now address \textbf{RQ2}. To test the robustness of our system to query variations, we created alternative phrasings for 18 queries from AskNature, resulting in 18 pairs of queries. 
We evaluated these pairs using two metrics: First, we looked at the number of shared results among the top 15 retrieved sentences for each pair. Second, we used the Rank-biased Overlap (RBO)~\cite{rbo}, which compares two ranked lists, taking into account both the shared items and their order, providing a score between 0 and 1. 

\xhdr{Results}
On the full dataset, \BARcode averages an RBO of 0.13 and of 2.83 shared items. This can be attributed to the size of the dataset it operates on and the number of possible good answers. On the filtered data, the average is RBO of 0.31 and 6.8 shared items (hinting at similar items ranked at different places). In other words, the system is more robust on the filtered data.

\subsection{Evaluating Bio-inspiration Scores}
\label{sec:exp1}

We now aim to assess if the filtering module (Section \ref{sec:filter}) identifies the kind of sentences we are after (\textbf{RQ3}). 
We compare to Cheong's rule-based approach \cite{Cheong2014}, the previous state-of-the-art. 

We lack a database of gold-standard labeled candidates, and the majority of sentences in our corpus are unlikely to contain challenge-strategy pairs. Thus, we randomly selected 4 verbs from the titles of AskNature articles. We extracted all sentences containing those verbs from Wikipedia and computed their ranking according to Cheong and according to our method. We sampled ten sentences per verb from the top and the bottom of each ranked list, yielding a total of 160 sentences for evaluation. We evaluate the bottom of the lists as well to provide a rough estimate of recall.

\begin{table}[t!]
  \centering
  \begin{tabular}{l p{2cm} c c p{1cm}}
    \toprule
    \textbf{Top K} & \textbf{Calculation type} & \textbf{\BARcode} & \textbf{Cheong} \\
    \midrule
    \multirow{2}{*}{High} & Strict & \textbf{0.56} & 0.22 \\
    & Liberal & \textbf{0.80} & 0.25 \\ \hline
    \multirow{2}{*}{Low} & Strict & 0.13 & \textbf{0.15} \\
    & Liberal & 0.28 & \textbf{0.33} \\
    \bottomrule
  \end{tabular}
  \caption{Comparing the performance of \BARcode and Cheong's rule-based approach. Our approach outperforms Cheong's method in top 10 score precision for both strict and liberal cutoffs. Recall is similar, with our approach having slightly fewer inspirational sentences in the bottom 10.}
  \label{tab:eval1_results}
\end{table}

Four members of the research team, blind to condition, scored the candidates on a scale of 0 (definitely does not contain a strategy to address a challenge) to 2 (definitely contains a strategy to address a challenge). 

Inter-rater reliability was acceptable (average pairwise Spearman's $\rho$ = .60). We computed a score for each match by averaging the 4 ratings, at both strict (average = 2) and liberal (average \(\ge\) 1) cutoffs.

\xhdr{Results}
Table \ref{tab:eval1_results} demonstrates that our approach substantially outperforms Cheong's rule-based approach in terms of precision for both strict (0.56 vs. 0.22) and liberal (0.80 vs. 0.25) cutoffs in the top 10. Our approach exhibits marginally fewer inspirations in the bottom 10 compared to the rule-based method, hinting that it does not compromise recall.

\section{Ideation Experiment : Case Studies}
\label{sec:ideation}
To evaluate whether our results extend to real-world scenarios, we recruited three experts engaged in innovative bioengineering projects:  
E1 sought inspiration for maintaining a stable temperature of liquid inside a tube, 
E2 wished to measure metabolic activity and cell assembly, and E3 sought new ways to record electrical activity from cell groups.

We guided the experts to formulate queries in a ``verb-object'' format, and instructed them to abstract their needs (rather than use extremely specific technical terms from their domain). We used the sped-up version of our algorithm to allow live interaction with the search engine.

\begin{table}[t!]
  \centering
  \begin{tabular}{c l c c c }
    \toprule
    \textbf{} & \textbf{Queries} & \textbf{\#} &  \textbf{Known} & \textbf{Interest}  \\
    \midrule
    \multirow{6}{*}{E1} & maintain aqueous  & 30 & 4 & 2 (1)\\ 
    & temperature & & &  \\ \cline{2-5}
    & regulate temperature & 15 & 0 & 1 (0) \\ \cline{2-5}
    & provide insulation in  & 15 & 0 & 1 (0) \\ 
    & extreme temperatures & & &  \\ \cline{2-5}
    & conserve heat & 15 & 0 & 1 (0) \\ \hline
    \multirow{4}{*}{E2} & measure metabolic & 25 & 1 & 6 (1) \\
    & activity & & &  \\ \cline{2-5}
    & sense lactate & 15 & 0 & 3 (2) \\ \cline{2-5} 
    & assemble cells & 15 & 2 & 3 (2) \\ \hline
    \multirow{4}{*}{E3} & sense electrical fields & 25 & 0 & 10 (3) \\ \cline{2-5}
    & recognize electrical  & 15 & 0 & 2 (2) \\ 
    & activity in a specific & & &  \\
    & radius & & &  \\
    \bottomrule
  \end{tabular}
  \caption{Ideation experiment. For each expert query, we show the number of results, number of results they were already familiar with (and were valuable), number of results they found somewhat interesting or interesting (score 1 or 2). In parenthesis, the number of very interesting results (score 2). 
  33\% of the relevant results (known or interesting) were considered highly interesting, and 21\% of them were already known to the experts.
  } 
  \label{tab:case_study_queries}
\end{table}

\xhdr{Known Directions}
The experts were pleasantly surprised
by the search results \BARcode returned and noted several that
matched their own findings. : E1 mentioned, ``...this would be the best solution for us... [wrapping their tube with another tube and maintaining a high level of moisture between them] ...we already thought about this idea, it is similar to the process used to pasteurize milk and heat blood samples''. Similarly, E2 stated, ``It is something that I already know and try to apply...''. 

Overall, 21\% of the relevant results were familiar to the experts (see Table \ref{tab:case_study_queries}). We take this as a positive signal, indicating the \BARcode does indeed manage to identify directions that experts had independently considered.

\xhdr{Novel and Inspirational}
\BARcode inspired experts to generate diverse ideas, revealing new perspectives and insights. For instance, E2 expressed enthusiasm, stating, ``this is the best result so far...it (catfish) has a system through its facial tasting sensors can detect metabolites... and I am looking for sensors for metabolites... I am intrigued to know how this mechanism works... also this mechanism seems to be constantly active, unlike current techniques''. Regarding the query of ``measure metabolic activity'', E2 stated ``...this is very interesting... I did not expect this result... it mentions that a low metabolic rate reduces mortality and increases longevity... I would be interested in exploring this''. In another instance, E1 dwelled on a moth's insulation layer used to prevent overheating. E1 highlighted not having considered such a structure before, and that it could serve as a more effective insulation layer.  
Experts rated each result on a scale from 0 (not interesting) to 2 (very interesting). 

Table \ref{tab:case_study_queries} shows the search queries, number of \BARcode results read by the expert during the session, and number of entries found interesting by experts (in parenthesis: number of very interesting results). We note that all but three of the queries led to at least one inspiration deemed ``very interesting'' by the expert.

After the interview, we thanked the experts for their participation. No other compensation
was provided.

\section{Conclusions and Future Work}
Biologically Inspired Design (BID), or Biomimicry, is a problem-solving methodology that applies analogies from nature to solve challenges. Finding relevant biological solutions for real-world problems poses significant challenges, both due to the limited biological knowledge engineers and designers typically possess and the scarcity of sources of inspiration. Current BID datasets are manually curated, and scaling them up requires expensive manual effort.

This paper introduces \BARcode (Biological Analogies Retriever), a flexible search engine that automatically mines bio-inspirations from the web. Leveraging advancements in natural language understanding and data programming, \BARcode identifies potential solutions rooted in nature for various engineering challenges. Our experiments confirm \BARcode's ability to extract valuable inspiration from large corpora, including famous historical examples of biomimicry. \BARcode was able to find promising inspirations for experts in engineering for their research problems.

We note that across our piloting of the system, we noticed substantial variance in performance across queries. Some of this might be due to the idiosyncrasies of the composition of our corpus; however, our experiences with the case study experts, in particular, suggest that exploring interactive mechanisms for query reformulation and expansion (perhaps with the help of large language models) might significantly improve the user experience.

In the future, we hope to apply \BARcode to different, more detailed sources of biological data, such as scientific papers. In this work, we have chosen to concentrate on retrieving inspirations; an exciting direction for future exploration is integrating \BARcode with generative artificial intelligence technologies to generate a sketch of the solution, adapting the biological strategy retrieved by \BARcode to the user's problem.
We see \BARcode as a step towards addressing the challenges that have historically hindered the practical application of BID to engineering innovation.

\section*{Acknowledgements}
We would like to thank Hyadata Lab and NLP-HUJI members for their thoughtful remarks, and all the participants in our experiments for their efforts. We also thank the reviewers for their insightful comments. This work was supported by the European Research Council (ERC) under the European Union’s Horizon 2020 research and innovation program (grant no. 852686, SIAM), Office of Naval Research, and Carnegie Mellon Center for Knowledge Acceleration. Hen Emuna was supported by Sam and Betty Sandler and the Samuel and Lottie Rudin Scholarship.

\bibliography{main}

\newpage
\appendix
\section{Appendix}

\subsection{Calculating Relevance Score}
\label{sec:relevance_app}

\subsubsection{QA-SRL}
Here, we demonstrate the use of QA-SRL to extract relevant candidate phrases and discuss its limitations. An example of sentence parsing using QA-SRL is shown in Table \ref{tab:qasrl}.

We omitted questions like ``when'' or ``who'' and those not ending with a verb, as they are unlikely to offer valuable inspiration. For instance, consider the sentence related to the shark in Table \ref{tab:qasrl}: [Question]: ``Who detects something?'' [Answer]: ``them''. 
Turning to the sentence involving the Peregrine falcon from Table \ref{tab:filtering_score}, it illustrates how questions not ending with a verb tend to be irrelevant. For example, [Question]: ``What reduces something?'' [Answer]: ``the bird''. In contrast, questions ending with a verb are more likely to pinpoint challenge-related phrases, as seen with [Question]: ``What is being reduced?'' [Answer]: ``the change in air pressure''.

\xhdr{Method Limitations}
Long, convoluted sentences processed by QA-SRL can yield verbose candidate phrases, which can potentially lead to missing relevant sentences. As shown in Table \ref{tab:qasrl}, the verb ``allows'', extracts the candidate ``allow to detect electricity emitted by other animals''. In contrast, the Dependency parsing method extracts a more concise phrase (i.e., ``detect electricity''), achieving a higher cosine similarity score for relevant queries, such as ``sense electrical activity'' (0.83 vs. 0.65). 

In certain cases, QA-SRL fails to extract the relevant candidate phrase, as seen in the Pelican sentence in Table \ref{tab:runing_example}. For the verb ``keep'', it extracts only the phrase ``keep pelican'', which does not effectively address any challenge.

\begin{table*}[t!]
\centering
\resizebox{\linewidth}{!}{%
\begin{tabular}{m{0.077\linewidth} m{0.233\linewidth} m{0.296\linewidth} m{0.331\linewidth}} 
\toprule
\multicolumn{4}{m{0.937\linewidth}}{\textit{\textbf{Input sentence}:} However, as with most other sharks, including other members of the family Scyliorhinidae, they are believed to have a well-developed sense of smell, and are electroreceptive, which allows them to detect electricity emitted by other animals, and may also allow them to detect magnetic fields, which aids in navigation.} \\
\midrule
\textbf{Verb} & \textbf{Question} & \textbf{Answer} & \textbf{Candidate Phrase} \\
\midrule
emitted & what is emitted? & electricity & emit electricity \\
\midrule
allows & what does something allow? & to detect electricity emitted by other animals & allow to detect electricity emitted by other animals \\
\midrule
detect & what is being detected? & magnetic fields & detect magnetic fields \\
       & what does something detect? & electricity emitted by other animals & detect electricity emitted by other animals \\
\midrule
aids & what does something aid? & in navigation & aid in navigation \\
\bottomrule
\end{tabular}
}
\captionsetup{justification=raggedright,singlelinecheck=false}
\caption{Illustrative Example: Extracting Candidate Phrases from Complex Sentences using QA-SRL.}
\label{tab:qasrl}
\end{table*}

\subsubsection{Generating Patterns}
Here, we describe how we used spaCy's DependencyMatcher to create patterns for extracting candidate phrases (i.e., verb-object pairs) from each sentence (Section \ref{sec:match}).

The DependencyMatcher is a rule-based matcher that enables matching on dependency trees using Semgrex operators based on rules describing token attributes. Rules can refer to token annotations (such as part-of-speech tags), as well as lexical attributes like the token’s dependency label. The operators are used to define the relationship between a headword and its dependent. For instance, the rule 'A $<$ B' indicates that A is the immediate dependent of B.
The keys used to define the patterns are detailed in Table \ref{tab:keys}.

Specifically, we defined 10 patterns in which in all of them the head token is always a verb, and its dependent is mostly a direct object. However, patterns can vary in complexity, ranging from simple ones like a verb and its direct object to more intricate configurations where a verb is the head of a direct object with an auxiliary verb, and the direct object has additional dependents like modifiers. Table \ref{tab:dependency_patterns} presents all 10 patterns, each described using DependencyMatcher's keys as shown in Table \ref{tab:keys}.

\begin{table}[t!]
\centering
  \begin{tabular}{ll}
    \toprule
    \textbf{NAME} & \textbf{DESCRIPTION} \\
    \midrule
    LEFT\_ID & The name of the left-hand node in \\ & the relation, which has been defined \\ & in an earlier node (Type: str). \\
    REL\_OP & An operator that describes how the \\ & two nodes are related (Type: str). \\
    RIGHT\_ID & A unique name for the right-hand \\ & node in the relation (Type: str). \\
    RIGHT\_ATTRS & The token attributes match for \\ & the right-hand  node in the same \\ & format as patterns provided to the \\ & regular token-based (Type: str). \\
    \bottomrule
  \end{tabular}
  \caption{Keys used in spaCy's DependencyMatcher Pattern Format: Derived from spaCy's API.}
  \label{tab:keys}
\end{table}

\begin{table*}[t!]
  \centering
  \begin{tabular}{c l}
    \toprule
    \textbf{\#} & \textbf{Pattern Description} \\
    \midrule
    \multirow{3}{*}{1} &
    \parbox[t]{16cm}{\texttt{\{ "RIGHT\_ID": "verb", 
    "RIGHT\_ATTRS": \{"POS": "VERB"\} \}, \\
    \{ "LEFT\_ID": "verb", 
    "REL\_OP": "$>$", 
    "RIGHT\_ID": "object", 
    "RIGHT\_ATTRS": \{"DEP": "dobj", "POS": "NOUN"\} \}
    }} \\
    \midrule
    \multirow{5}{*}{2} &
    \parbox[t]{16cm}{\texttt{\{ "RIGHT\_ID": "verb",
    "RIGHT\_ATTRS": \{"POS": "VERB"\} \}, \\
    \{ "LEFT\_ID": "verb", 
    "REL\_OP": "$>$", 
    "RIGHT\_ID": "object", 
    "RIGHT\_ATTRS": \{"DEP": "dobj"\} \}, \\
    \{ "LEFT\_ID": "object", 
    "REL\_OP": "$>$", 
    "RIGHT\_ID": "mod/comp", 
    "RIGHT\_ATTRS": \{"DEP": \{"IN": ["amod", "compound"]\}\} \}
    }}
    \\
    \midrule
    \multirow{5}{*}{3} &
    \parbox[t]{16cm}{\texttt{\{ "RIGHT\_ID": "verb", 
    "RIGHT\_ATTRS": \{"POS": "VERB"\} \}, \\
    \{ "LEFT\_ID": "verb", 
    "REL\_OP": "$>$", 
    "RIGHT\_ID": "object", 
    "RIGHT\_ATTRS": \{"DEP": "dobj"\} \}, \\
    \{ "LEFT\_ID": "object", 
    "REL\_OP": "$<$", 
    "RIGHT\_ID": "mod/comp", 
    "RIGHT\_ATTRS": \{"DEP": \{"IN": ["amod", "compound"]\}\} \}
    }}
    \\
    \midrule
    \multirow{7}{*}{4} &
    \parbox[t]{16cm}{\texttt{\{ "RIGHT\_ID": "verb", 
    "RIGHT\_ATTRS": \{"POS": "VERB"\} \}, \\
    \{ "LEFT\_ID": "verb", \
    "REL\_OP": "$>$", 
    "RIGHT\_ID": "object", 
    "RIGHT\_ATTRS": \{"DEP": "dobj"\} \}, \\
    \{ "LEFT\_ID": "object", 
    "REL\_OP": "$>$", 
    "RIGHT\_ID": "preposition",
    "RIGHT\_ATTRS": \{"DEP": \{"IN": ["prep", "xcomp"]\}\} \}, \\
    \{ "LEFT\_ID": "preposition", 
    "REL\_OP": "$>$", 
    "RIGHT\_ID": "pobj", 
    "RIGHT\_ATTRS": \{"DEP": "pobj"\} \} 
    }}
    \\
    \midrule
    \multirow{5}{*}{5} &
    \parbox[t]{16cm}{\texttt{\{ "RIGHT\_ID": "verb", 
    "RIGHT\_ATTRS": \{"POS": "VERB"\} \}, \\
    \{ "LEFT\_ID": "verb", 
    "REL\_OP": "$>$", 
    "RIGHT\_ID": "object", 
    "RIGHT\_ATTRS": \{"DEP": "dobj"\} \}, \\
    \{ "LEFT\_ID": "verb", 
    "REL\_OP": ".", 
    "RIGHT\_ID": "auxiliary verb", 
    "RIGHT\_ATTRS": \{"DEP": \{"IN": ["aux", "xcomp"]\}, "POS": "VERB"\} \} 
    }}
    \\
    \midrule
    \multirow{3}{*}{6} &
    \parbox[t]{16cm}{\texttt{\{ "RIGHT\_ID": "verb", 
    "RIGHT\_ATTRS": \{"POS": "VERB"\} \}, \\
    \{ "LEFT\_ID": "verb", 
    "REL\_OP": "$>$", 
    "RIGHT\_ID": "object predicate", 
    "RIGHT\_ATTRS": \{"DEP": \{"IN": ["oprd", "acomp", "prt"]\}\} \}
    }}
    \\
    \midrule
    \multirow{5}{*}{7} &
    \parbox[t]{16cm}{\texttt{\{ "RIGHT\_ID": "verb", 
    "RIGHT\_ATTRS": \{"POS": "VERB"\} \}, \\
    \{ "LEFT\_ID": "verb", 
    "REL\_OP": "$>$", 
    "RIGHT\_ID": "preposition", 
    "RIGHT\_ATTRS": \{"DEP": "prep"\} \}, \\
    \{ "LEFT\_ID": "preposition", 
    "REL\_OP": "$>$", 
    "RIGHT\_ID": "object of a preposition", 
    "RIGHT\_ATTRS": \{"DEP": "pobj", "POS": "NOUN"\} \} 
    }}
    \\
    \midrule
    \multirow{3}{*}{8} &
    \parbox[t]{16cm}{\texttt{\{ "RIGHT\_ID": "verb", 
    "RIGHT\_ATTRS": \{"POS": "VERB"\} \}, \\
    \{ "LEFT\_ID": "verb", 
    "REL\_OP": "$>$", 
    "RIGHT\_ID": "noun phrase as adverbial modifier", 
    "RIGHT\_ATTRS": \{"DEP": "npadvmod"\} \} 
    }}
    \\
    \midrule
    \multirow{3}{*}{9} &
    \parbox[t]{16cm}{\texttt{\{ "RIGHT\_ID": "verb", 
    "RIGHT\_ATTRS": \{"POS": "VERB"\} \}, \\
    \{ "LEFT\_ID": "verb", 
    "REL\_OP": ".", 
    "RIGHT\_ID": "adposition", 
    "RIGHT\_ATTRS": \{"DEP": "prt", "POS": "ADP"\} \} 
    }}
    \\
    \midrule
    \multirow{3}{*}{10} &
    \parbox[t]{16cm}{\texttt{\{ "RIGHT\_ID": "verb", 
    "RIGHT\_ATTRS": \{"POS": "VERB"\} \}, \\
    \{ "LEFT\_ID": "verb", 
    "REL\_OP": "$>$", 
    "RIGHT\_ID": "nominal subject (passive)", 
    "RIGHT\_ATTRS": \{"DEP": "nsubjpass", "POS": "NOUN"\} \} 
    }}
    \\
    \bottomrule
\end{tabular}
\caption{Dependency Parsing Patterns for Candidate Phrase Extraction: Table indicating patterns for extracting phrases from sentences, with "\#" denoting the pattern number.}
\label{tab:dependency_patterns}
\end{table*}

\subsubsection{Creating Weighted Score}
Here, we explain how we used a simple classifier to combine the four scores from SBERT and NLI (cosine, entailment, neutral, and contradiction) into a single weighted score.

In a preliminary experiment, we collected 48 queries from AskNature's search results (different from those used in this paper). We selected the top 15 phrases with the highest cosine similarity score, the top 15 with the highest entailment score, and an additional 15 phrases with the lowest contradiction score to balance the training data. This process resulted in a total of 1005 pairs after removing duplicates. An expert member of our team labeled these pairs as relevant or irrelevant based on their semantic relevance.

Using these labels, we trained an SVM to classify query-candidate phrase pairs as either relevant or irrelevant based on the four associated scores (the fraction of relevant labels is 63\%). To determine optimal hyperparameters, we conducted a grid search (see detailed information below). Our primary goal was to maximize accuracy while maintaining a careful balance between precision and recall. Put simply, given a query, the matching function applies these weights to the four scores of each candidate phrase, generating a single score indicating its relevance to the query.

\xhdr{Grid Search}
To identify optimal hyperparameters, we conducted a grid search using Scikit-Learn's SVM model. A consistent random seed was applied for data splitting across runs, with a test size of 0.3. The parameters considered included various kernel types ('linear', 'RBF', 'sigmoid', 'poly'), C values (0.1, 1, 10, 100, 1000), gamma values (0.1, 0.01, 0.001, 0.0001), and degrees (2, 3, 4) for the polynomial kernel. This resulted in a total of 650 fits, using 5-fold cross-validation for each of the 130 candidates. The average fit time was 16.56 seconds. We optimized precision to 0.83 with the selected hyperparameters: C=100, degree=2, gamma=0.1, and kernel='poly'.

The grid search, algorithm runs, experiments, and analysis were conducted on a system with the following specifications: Intel(R) Core(TM) i7-8550U CPU @ 1.80GHz, 16.0 GB RAM, running Microsoft Windows 10 Pro.

\subsection{Data Filtering}
\label{sec:filtering_app}

\subsubsection{Snorkel's Generative Model}
We trained Snorkel's generative model with a learning rate of $0.0001$ for $3000$ epochs. All other settings were left at their default values.

\subsection{Evaluation}
\label{sec:eval_app}

\subsubsection{Queries}
See Table \ref{tab:asknature_phrases} for a full list of queries used in our experiments.

\begin{table}[hbt]
\centering
\begin{tabular}{p{3.5cm}p{3.1cm}}
\toprule
\textbf{AskNature} & \textbf{Paraphrases} \\
\midrule
absorb carbon dioxide & capture carbon \\ & dioxide \\
repel water & prevent water \\ & absorption \\
renew cell & regenerate cell \\
produce color & create pigment \\
produce sound & generate sound \\
change form & modify shape \\
blend into environment & use camouflage \\
allow flexibility & provide elasticity \\
control gliding & maneuver in air \\
provide lift & provide lifting force \\
move efficiently through water & reduce water resistance \\
reduce drag & minimize friction \\
allow floating & enable buoyancy \\
sense vibration & detect vibration \\
enhance night vision & see in the dark \\
improve vision & enhance vision \\
enable clear vision in water & improve underwater vision \\
sense electrical energy & detect electric field \\
store liquid & - \\
produce electricity & - \\
change color & - \\
prevent slipping & - \\
stay underwater & - \\
control buoyancy & - \\
\bottomrule
\end{tabular}
\caption{Original and Paraphrased AskNature Queries: AskNature queries and their corresponding paraphrased versions, along with the inclusion of six additional queries lacking corresponding paraphrases.}
\label{tab:asknature_phrases}
\end{table}

\subsubsection{Annotation Task}
\begin{figure*}[t!]
  \includegraphics[width=0.9\textwidth]{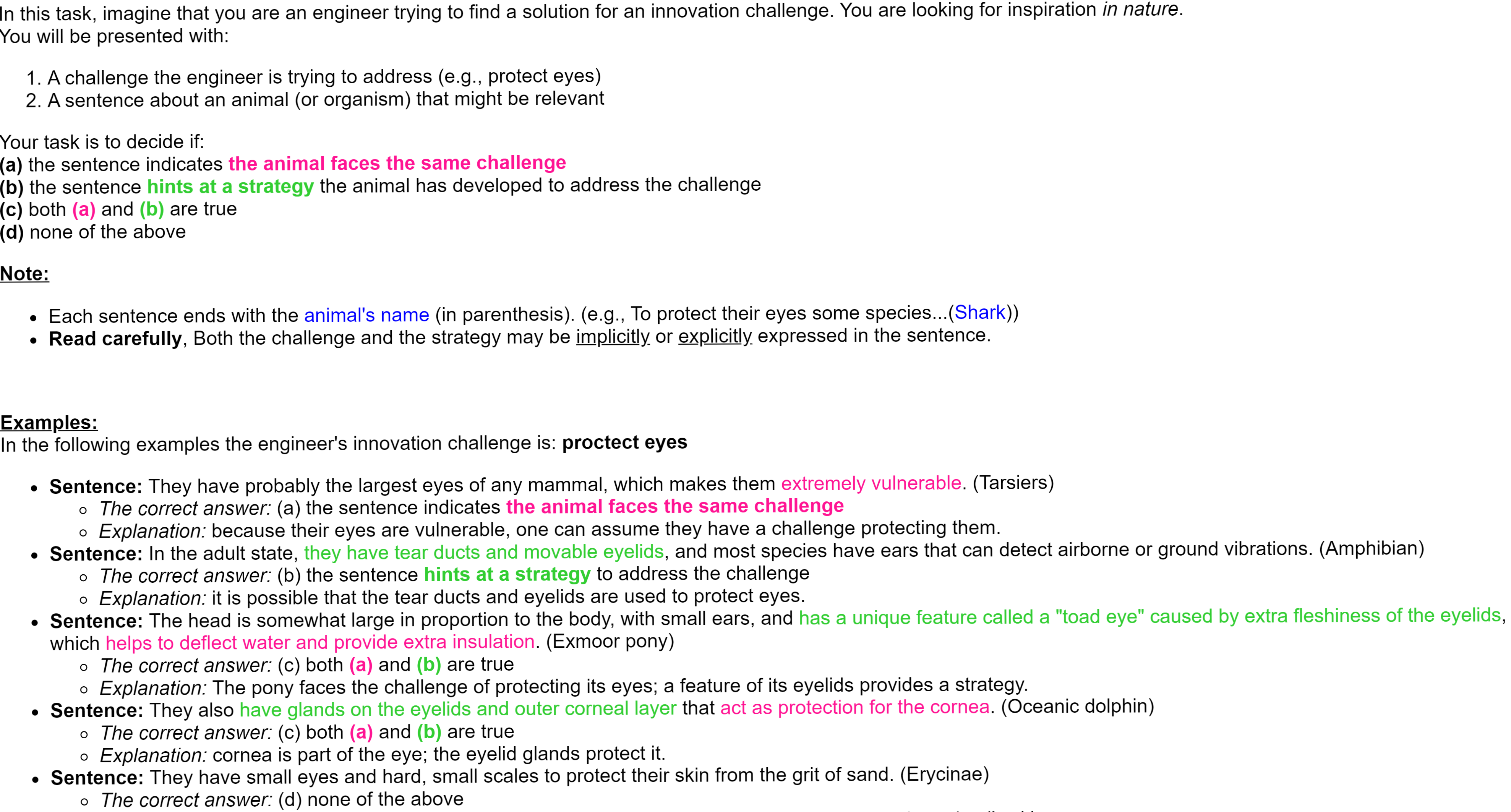}
  \caption{Annotation Instructions for Amazon Mechanical Turk Task: Guidelines covering task framework, potential difficulties, detailed examples, and rationale for accurate annotations.}
  \label{fig:task_instructions}
\end{figure*}

\begin{figure*}[t!]
  \includegraphics[width=0.9\textwidth]{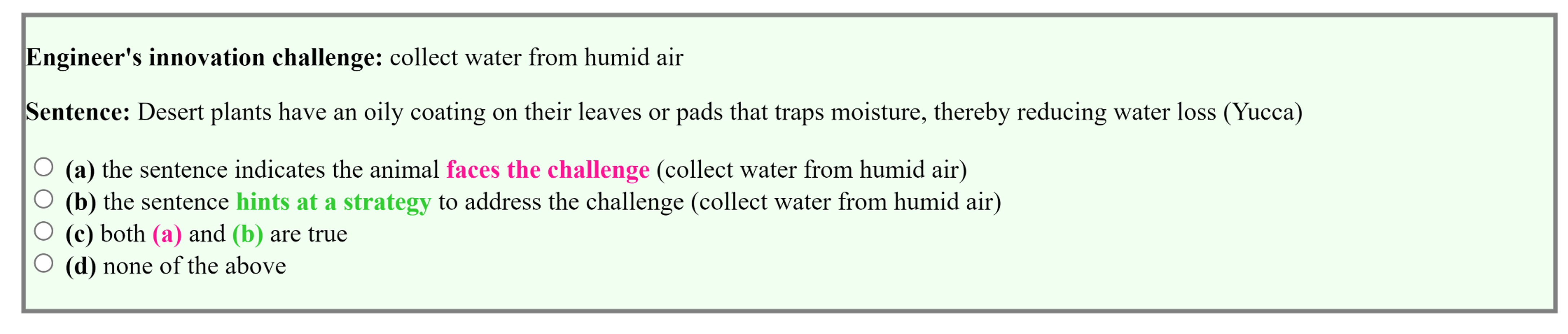}
\caption{Annotation Example Task: A snapshot of an annotation task performed by Amazon Mechanical Turk workers.}
\label{fig:task_example}
\end{figure*}

Annotations begin with comprehensive task instructions shown in Figure \ref{fig:task_instructions}. These guidelines cover the task approach, highlight potential challenges, and present five examples addressing each of the four optional answers. Each example is followed by an explanation of the rationale behind the correct answer. After reading the instructions, workers take a qualification test, wherein they must accurately annotate three tasks without errors. Those who pass the test proceed to the annotation task, illustrated in Figure \ref{fig:task_example}. 

\subsubsection{Famous Biomimicry Examples}
Here are the sentences that inspired famous historical examples of biomimicry retrieved by our system:

\begin{compactitem}
    \item Namibian desert beetles inspired the idea of capturing cooling tower plumes to reduce water usage. The beetle's behavior is described as follows: ``Facing into the breeze, with its body angled at 45 degrees, the beetle catches fog droplets on its hardened wings''. This example ranked 8 for the query ``collect water from humid air''. Another sentence about the same organism ranked 21 for the same query: ``Droplets flatten as they make contact with the hydrophilic surfaces, preventing them from being blown by wind and providing a surface for other droplets to attach''. 
    \item Gecko feet inspired the development of residue-free adhesion technology. The effect of humidity on gecko adhesion is discussed: ``Increasing humidity typically fortifies gecko adhesion, even on hydrophobic surfaces, yet is reduced if completely immersed in water''. Ranked 9 for the query ``provide adhesion''.
    \item Shark skin inspired the creation of Speedo's swimsuit designed for faster swimming. The advantage of shark skin's dermal teeth is highlighted: ``Their dermal teeth give them hydrodynamic advantages as they reduce turbulence when swimming''. Ranked 10 for the query ``reduce fluid drag''.
    \item Termite mounds inspired the concept of passively cooled buildings. The role of mounds' architecture in enhancing air circulation is mentioned: ``Wind blowing across the tops of the towers enhances the circulation of air through the mounds, which also include side vents in their construction''. Ranked 14 for the query ``provide self-regulating ventilation system''.
\end{compactitem}

\end{document}